\documentclass[11pt,a4paper]{article}
\usepackage[hyperref]{emnlp2018}
\usepackage{times}
\usepackage{latexsym}
\usepackage{amsmath}
\usepackage{xcolor}
\usepackage{booktabs}
\usepackage{url}
\usepackage{bm}
\usepackage{array}
\usepackage{graphicx}
\usepackage{graphics}

\usepackage{arabtex}
\usepackage[T1]{fontenc}
\newcommand{\hide}[1]{}

\aclfinalcopy 

\let\vec\mathbf
\newcolumntype{C}[1]{>{\centering\let\newline\\\arraybackslash\hspace{0pt}}m{#1}}


\title{Utilizing Character and Word Embeddings for  Text Normalization\\ 
with Sequence-to-Sequence Models}

\author{Daniel Watson, Nasser Zalmout and Nizar Habash \\
Computational Approaches to Modeling Language Lab\\
   New York University Abu Dhabi \\
   {\tt\{daniel.watson,nasser.zalmout,nizar.habash\}@nyu.edu}}

\date{}

\begin{document}
\maketitle

\setarab
\novocalize

\begin{abstract}
Text normalization is an important enabling technology for several NLP tasks. Recently, neural-network-based approaches have outperformed well-established models in this task. However, in languages other than English, there has been little exploration in this direction. Both the scarcity of annotated data and the complexity of the language increase the difficulty of the problem. To address these challenges, we use a sequence-to-sequence model with character-based attention, which in addition to its self-learned character embeddings, uses word embeddings pre-trained with an approach that also models subword information. This provides the neural model with access to more linguistic information especially suitable for text normalization, without large parallel corpora. We show that providing the model with word-level features bridges the gap for the neural network approach to achieve a state-of-the-art $F_1$ score on a standard Arabic language correction shared task dataset.

\end{list}
\end{abstract}

\section{Introduction}

Text normalization systems have many potential applications -- from  assisting native speakers and language learners with their writing, to supporting NLP applications with sparsity reduction by cleaning large textual corpora. This can help improve benchmarks across many NLP tasks.

In recent years, neural encoder-decoder models have shown promising results in language tasks like translation, part-of-speech tagging, and text normalization, especially with the use of an attention mechanism. In text normalization, however, previous state-of-the-art results rely on developing many other pipelines on top of the neural model. Furthermore, such neural approaches have barely been explored for this task in Arabic, where previous state-of-the-art systems rely on combining various statistical and rule-based approaches.

We experiment with both character embeddings and pre-trained word embeddings, using several embedding models, and we achieve a state-of-the-art $F_1$ score on an Arabic spelling correction task.

\section{Related Work}
The encoder-decoder neural architecture \cite{DBLP:journals/corr/SutskeverVL14,cho-al-emnlp14} has shown promising results in text normalization tasks, particularly in character-level models \cite{Xie2016NeuralLC,ikeda2016japanese}. More recently, augmenting this neural architecture with the attention mechanism \cite{DBLP:journals/corr/BahdanauCB14,DBLP:journals/corr/LuongPM15} has dramatically increased the quality of results across most NLP tasks. However, in text normalization, state-of-the-art results involving attention (e.g., \citealt{Xie2016NeuralLC}) also rely on several other models during inference, such as language models and classifiers to filter suggested edits. Neural architectures at the word level inherently rely on multiple models to align and separately handle out-of-vocabulary (OOV) words \cite{yuan2016grammatical}.

In the context of Arabic, we are only aware of one attempt to use a neural model for end-to-end text normalization \cite{mastersthesis}, but it fails to beat all baselines reported later in this paper. Arabic diacritization, which can be considered forms of text normalization, has received a number of neural efforts
\cite{belinkov2015arabic,Abandah2015}.
However, state-of-the-art approaches for end-to-end text normalization rely on several additional models and rule-based approaches as hybrid models \cite{madamira,rozovskaya2014thecs,nawar2015wanlp,Nasser2017EMNLP}, which introduce direct human knowledge into the system, but are limited to correcting specific mistakes and rely on expert knowledge to be developed.

\section{Approach}

Many common mistakes addressed by text normalization occur fundamentally at the character level. Moreover, the input data tends to be too noisy for a word-level neural model to be an end-to-end solution due to the high number of OOV words. In Arabic, particularly, mistakes may range from simple orthographic errors (e.g., positioning of Hamzas) and keyboard errors to dialectal code switching and spelling variations, making the task more challenging than a generic language correction task. We opt for a character-level neural approach to capture these highly diverse mistakes. While this method is less parallelizable due to the long sequence lengths, it is still more efficient due to the small vocabulary size, making inference and beam search computationally feasible.

\subsection{Neural Network Architecture}

Given an input sentence $\vec{x}$ and its corrected version $\vec{y}$, the objective is to model $P(\vec{y}|\vec{x})$. The vocabulary can consist of any number of unique tokens, as long as the following are included: a padding token to make input batches have equal length, the two canonical start-of-sentence and end-of-sentence tokens of the encoder-decoder architecture, and an OOV token to replace any character outside the training data during inference. Each character $x_i$ in the source sentence $\vec{x}$ is mapped to the corresponding $d_{ce}$-dimensional row vector $\vec{c_i}$ of a learnable $d_{voc} \times d_{ce}$ embedding matrix, initialized with a random uniform distribution with mean 0 and variance 1.
  For the encoder, we learn $d$-dimensional representations for the sentence with two gated recurrent unit (GRU) layers \cite{cho-al-emnlp14}, making only the first layer bidirectional following \citet{45610}. Like long short-term memory \cite{Hochreiter:1997:LSM:1246443.1246450}, GRU layers are well-known to improve the performance of recurrent neural networks (RNN), but are slightly more computationally efficient than the former.

\begin{figure}[t!]
\centering
\includegraphics[width=0.4\textwidth]{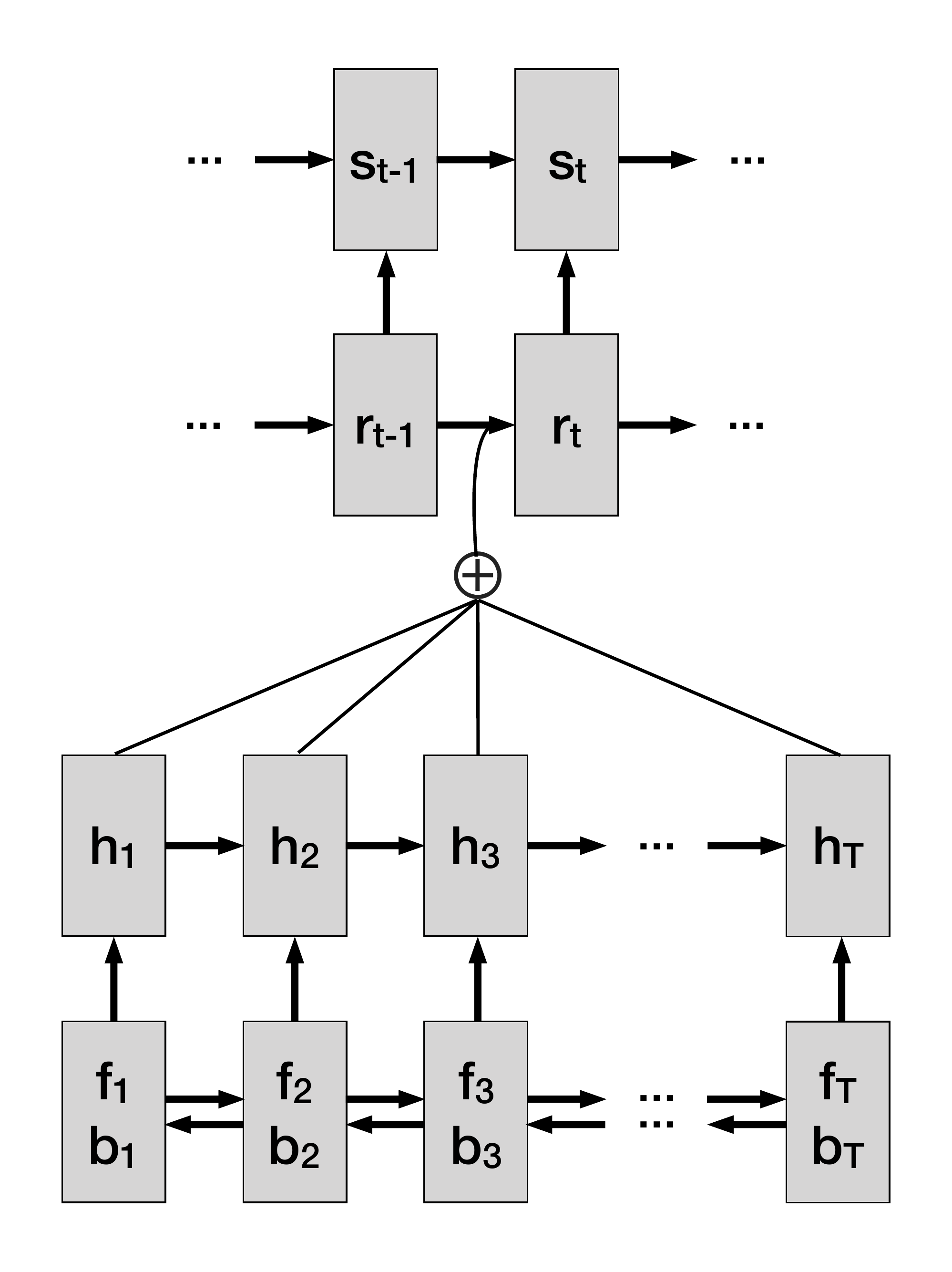}
\caption{Illustration of the encoder and decoder recurrent layers.}
\end{figure}

For the decoder, we use two GRU layers along with the attention mechanism proposed by \citet{DBLP:journals/corr/LuongPM15} over the encoder outputs $h_i$. The initial states for the decoder layers are learned with a fully-connected $\tanh$ layer in a similar fashion to \citet{cho-al-emnlp14}, but we do so from the first encoder output. During training, we use scheduled sampling \cite{DBLP:journals/corr/BengioVJS15} and feed the $d_{ce}$-dimensional character embeddings at every time step, but using a constant sampling probability. While tuning scheduled sampling, we found that introducing a sampling probability provided better results than relying on the ground truth, i.e., teacher forcing \cite{williams1989learning}. However, introducing a schedule did not yield any improvement as opposed to keeping the sampling probability constant and unnecessarily complicates hyperparameter search. For both the encoder and decoder RNNs, we also use dropout \cite{srivastava2014dropout} on the non-recurrent connections of both the encoder and decoder layers during training.

The decoder outputs are fed to a final softmax layer that reshapes the vectors to dimension $d_{voc}$ to yield an output sequence $\vec{y}$. The loss function is the canonical cross-entropy loss per time step averaged over the $y_i$.

\subsection{Word Embeddings}
To address the challenge posed by the small amount of training data, we propose adding pre-trained word-level information to each character embedding. To learn these word representations, we use FastText \cite{bojanowski2016enriching}, which extends Word2Vec \cite{mikolov2013distributed} by adding subword information to the word vector. This is very suitable for this task, not only because many mistakes occur at the character level, but also because FastText handles almost all OOVs by omitting the Word2Vec representation and simply using the subword-based representation. It is possible, yet extremely rare that FastText cannot handle a word-- this can occur if the word contains an OOV n-gram or character that did not appear in the data used to train the embeddings. It should also be noted that these features are only fed to the encoder layer; the decoder layer only receives $d_{ce}$-dimensional character embeddings as inputs, and the softmax layer has a $d_{voc}$-dimensional output.

Each character embedding $\vec{c_i}$ is replaced by the concatenation $\left[\vec{c_i}; \vec{w_j}\right]$ before being fed to the encoder-decoder model, where $\vec{w_j}$ is the $d_{we}$-dimensional word embedding for the word in which $\vec{c_i}$ appears in. This effectively handles almost all cases except white spaces, in which we just always append a $d_{we}$-dimensional vector $\vec{w_{\_}}$ initialized with a random uniform distribution of mean 0 and variance 1. For OOVs, we just append the whitespace embedding $\vec{w_{\_}}$ to the word's characters.

\begin{figure}[t!]
\centering
\includegraphics[width=0.4\textwidth]{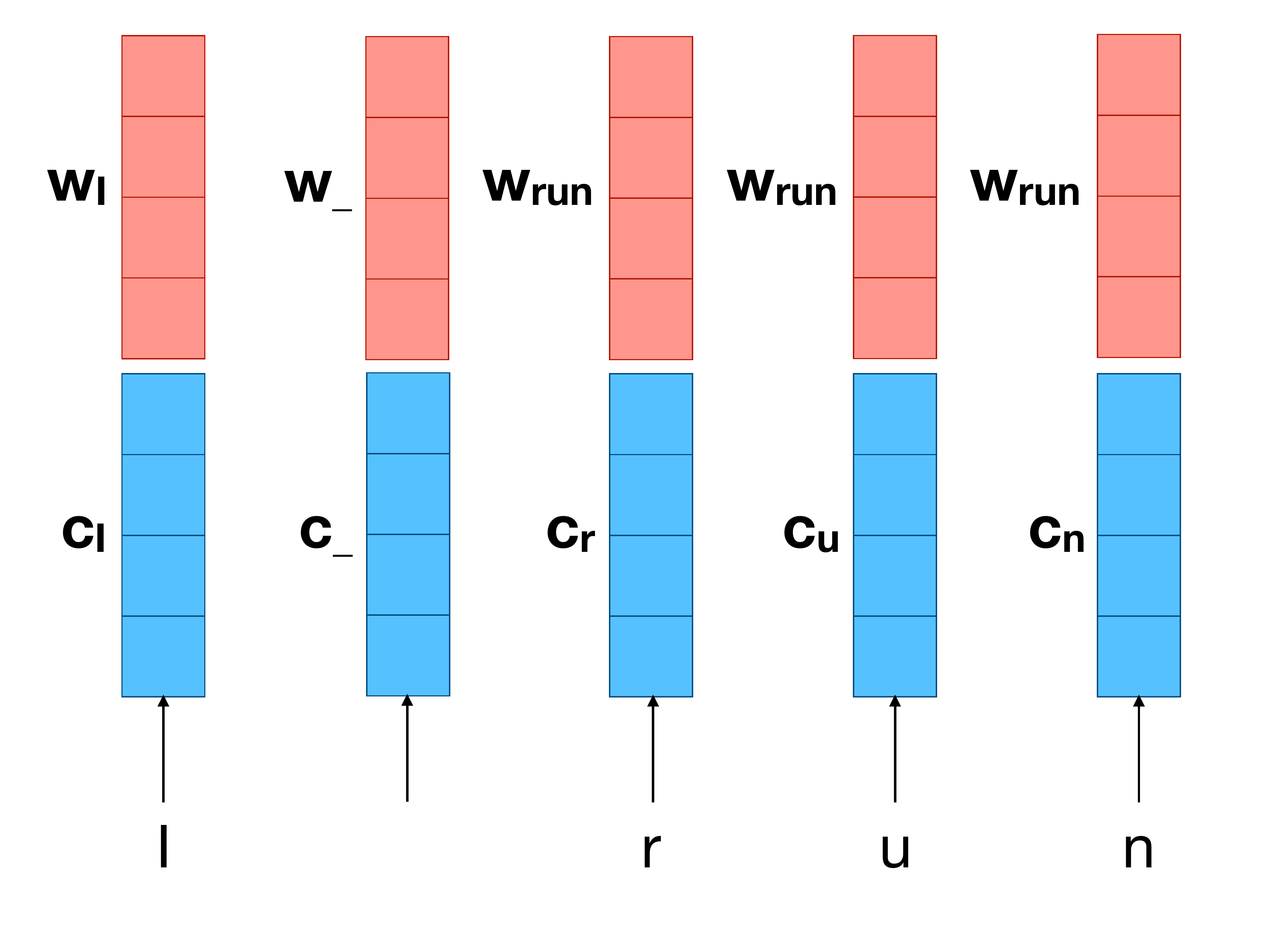}
\caption{Illustration showing how the character embeddings are enriched with word-level features.}
\end{figure}

\subsection{Inference}

During inference, the decoder layer uses a beam search, keeping a fixed number (i.e., beam width) of prediction candidates with the highest log-likelihood at each step. Whenever an "end-of-sentence" token is produced in a beam, the decoder stops predicting further tokens for it. We then pick the individual beam with the highest overall log-likelihood as our prediction. As a final step, we reduce text sequences that are repeated six or more times to a threshold of 5 repetitions (e.g., \verb|"abababababab"| $\rightarrow$  \verb|"ababababab"|). This helps address rare cases where the decoder misbehaves and produces non-stop repetitions of text, and also helps avoid extreme running times for the NUS MaxMatch scorer \cite{m2scorer}, which we use for evaluation and comparison purposes.

\section{Evaluation}

\subsection{Data}
We tested the proposed approach on the QALB dataset, a corpus for Arabic language correction and subject of two shared tasks \cite{zaghouani2014large,mohit2014first,rozovskaya2015second}. Following the guidelines of both shared tasks, we only used the training data  of the QALB 2014 shared task corpus (19,411 sentences). Similarly, the validation dataset used is only that of the QALB 2014 shared task, consisting of 1,017 sentences. We use two blind tests, one from each year. During training, we only kept sentences of up to 400 characters in length, resulting in the loss of 172 sentences.

\begin{table}[t!]
\centering
\scalebox{0.95}{
\begin{tabular}{lccc}
  \textbf{Baseline} & \bm{$P$} & \bm{$R$} & \bm{$F_1$} \\
  \hline
 MLE & 77.08 & 41.56 & 54.00 \\
  \hline
  MADAMIRA & 77.47 & 32.10 & 45.39 \\
  \hline
  MLE then MADAMIRA & 64.38 & 38.42 & 48.12 \\
  \hline
  MADAMIRA then MLE & 73.55 & 44.61 & \textbf{55.54} \\  \hline
\end{tabular}
}
\caption{Baselines scores on the QALB 2014 shared task development set.}
\label{table:1:baselines}
\end{table}

\subsection{Metric}
Like in the QALB shared tasks, we use the {\it MaxMatch} scorer to compute the optimal word-level edits that map each source sentence to its respective corrected sentence. We report the $F_1$ score of these edits against those provided in the gold data by the same tool. We compare against the best reported system in the QALB 2014 shared task test set (CLMB) \cite{rozovskaya2014thecs}, as well as the best in the QALB 2014 shared task development and the QALB 2015 shared task test sets (CUFE) \cite{nawar2015wanlp}.

\subsection{Baselines} We consider two different baselines. The first is the output from MADAMIRA \cite{madamira}, a tool for morphological analysis and disambiguation of Arabic. The second is using maximum likelihood estimation (MLE) based on the counts of the MaxMatch gold edits from the training data; that is, each word or phrase gets either replaced or kept as is, depending on the most common action in the training data. We found that, unlike \newcite{Eskander-Codafy:2013} suggested, first using MADAMIRA and then MLE yields better results than composing these in the reverse order. The results are presented in Table~\ref{table:1:baselines}.

\begin{table}[t]
\centering
\scalebox{1}{
\begin{tabular}{lccc}
  \textbf{Model} & \bm{$P$} & \bm{$R$} & \bm{$F_1$} \\
  \hline
  Wide embeds & 80.80 & 59.80 & 68.73 \\
  \quad + preprocessing & 79.63 & 58.81 & 67.57\\
  \hline
  Narrow embeds & 80.00 & 62.46 & \textbf{70.15} \\
  \quad + preprocessing & 80.25 & 57.80 & 67.20 \\
  \hline
  Concat embeds & 80.74 & 61.10 & 69.56 \\
  \quad + preprocessing & 79.81 & 58.28 & 67.37 \\
  \hline
  \hline
  CUFE \cite{nawar2015wanlp}  & -- & -- & 68.72 \\
  \hline
\end{tabular}
}
\caption{System scores on the QALB 2014 shared task development dataset for the different FastText embeddings.}
\label{table:2:dev}
\end{table}

\subsection{Model Settings}  In all experiments, we used batch and character embedding sizes of $b\!\!=\!\!d_{ce}\!\!=\!\!128$, hidden layer size of $d\!\!=\!\!256$, dropout probability of 0.1, decoder sampling probability of 0.35, and gradient clipping with a maximum norm of 10. When running all the trained models during inference, we used a beam width of 5. We used the Adam optimizer \cite{adam} with a learning rate of 0.0005, $\epsilon\!=\!\!1\!\cdot\!10^{-8}$, $\beta_1\!\!=\!\!0.9$, and $\beta_2\!\!=\!\!0.999$, and trained the model for 30 epochs. We report three different setups with FastText word embeddings: narrow, wide, and the concatenation of both. For each of these, we report results on two separately trained models: one without preprocessing, and one with MADAMIRA and then MLE preprocessing to the inputs. We also report an ablation study where we choose the best of these six trained models and compare against two separately trained models with identical setups, but using Word2Vec and no word-level features, respectively.

All the word embeddings used are of dimension $d_{we}\!\!\!=\!\!\!300$, and were trained with a single epoch over the Arabic Gigaword corpus \cite{LDC:Gigaword-5}. In the experiments including preprocessing, the respective word vectors were obtained from Gigaword preprocessed with MADAMIRA. The narrow and wide word embeddings were trained with context windows of sizes~2 and~5, respectively. All other hyperparameters were kept to the default FastText values, except the minimum n-gram size, which was reduced from~3 to~2 to compensate for single-character prefixes and suffixes that appear in Arabic when omitting the short vowels \cite{Erdmann-DAEmbeddings:2018}.

\begin{table}[t]
\centering
\scalebox{1}{
\begin{tabular}{lccc}
  \textbf{Model} & \bm{$P$} & \bm{$R$} & \bm{$F_1$} \\
  \hline
  No word embeds & 81.55 & 56.13 & 66.49 \\
  \hline
  Word2Vec & \textbf{82.16} & 51.53 & 63.33 \\
  \hline
  FastText & 80.00 & \textbf{62.46} & \textbf{70.15} \\
  \hline
\end{tabular}
}
\caption{Ablation tests on the QALB 2014 shared task development dataset. All settings used no preprocessing and narrow word embeddings.}
\label{table:3:ablation}
\end{table}

\subsection{Results}
Development set results are shown in Tables~\ref{table:2:dev} and~\ref{table:3:ablation}, test set results in Table~\ref{table:4:test}. In all models, training without preprocessing consistently yielded better results than their analogues with the inputs pre-fed to MADAMIRA and then MLE. All the FastText embeddings setups with no preprocessing outperformed the previous state-of-the-art results in the development dataset. We hypothesize that this is occurs because the model has access to more examples of errors to normalize during training, thereby increasing performance. The best performing model was that with the narrow word embeddings; consistent with the results of \citet{Zalmout-EGYMorphology:2018} showing the superior performance of narrow word embeddings over both wide embeddings and the concatenation of both.  This is justified by \citet{DBLP:journals/corr/Goldberg15c} and \citet{trask2015modeling}, who illustrate that wider word embeddings tend to capture more semantic information, while narrower word embeddings model more syntactic phenomena.

In our ablation study, we compared the performance of the narrow FastText embeddings against narrow Word2Vec embeddings trained over the same Arabic Gigaword corpus with the same hyperparameters, as well as to no word-level embeddings at all. The results, displayed in Table~\ref{table:3:ablation}, show that using only Word2Vec slightly increases precision but significantly hurts recall. This highlights the effectiveness of using FastText for text normalization, as well as the need to handle OOVs in a noisy context for word-level representations to help in this particular problem. Despite that having OOV cases can help the model by indicating that a word is likely erroneous, this does not provide linguistic information the way FastText does.
The narrow FastText embeddings with no preprocessing setup achieved state-of-the-art results in all three datasets, beating all systems in both the 2014 and 2015 QALB shared tasks in $F_1$ score. 

\subsection{Error Analysis}
We conducted a detailed error analysis of 101 sentences from the development set (6,370 words).
The sample contained 1,594 erroneous words (25\%). The errors were manually classified in a number of categories, which are presented in Table~\ref{table:errors}. 
The Table indicates the percentage of the error type in the whole set of errors as well as the error-specific {$F_1$} and an example.
Some common  problems,  Hamza  (glottal  stop)  and Ta Marbuta (feminine ending), are well handled in our best system. This is due to their commonality in the training data.  Other types are less common -- dialectal constructions, consonantal switches and Mood/Case.  Punctuation is very common, however it is also very idiosyncratic. We also note the presence of a small percentage~(0.5\%) of gold errors. For more information on Arabic language orthography issues from a computational perspective, see \citep{Buckwalter:2007, Habash_10_book, Habash_12_LREC}. 

\begin{table}[t]
\setlength{\tabcolsep}{2pt}
\centering
\scalebox{1}{
\begin{tabular}{lcc}
  &   (2014) & (2015) \\
  \textbf{Model} & \bm{$F_1$} & \bm{$F_1$}   \\
  \hline
   MADAMIRA then MLE & 55.56 & 60.98 \\
  \hline
  CLMB \cite{rozovskaya2014thecs} & 67.91 & -- \\
  \hline
  CUFE \cite{nawar2015wanlp} & 66.68 & 72.87 \\
  \hline
  \hline
  Narrow embeds & \textbf{70.39} & \textbf{73.19} \\  \hline
\end{tabular}
}
\caption{System score on the QALB 2014 and QALB 2015 shared task test datasets.}
\label{table:4:test}
\end{table}

\section{Conclusion and Future Work}

We propose a novel approach to text normalization by enhancing character embeddings with word-level features that model subword information and model syntactic phenomena. This significantly improves the neural model's recall, allowing the correction of more complex and diverse errors. Our approach achieves state-of-the-art results in the QALB dataset, despite it being small and seemingly unsuited for a neural model. Moroever, our neural model is sophisticated enough to not benefit from preprocessing techniques that reduce the number of errors in the data. Our approach is general enough to be implemented for any other text normalization task and does not rely on domain-specific knowledge to develop. 

Future directions include expanding the number of training pairs via synthetic data generation, where generative models can potentially add human-like errors to a large, unannotated corpus. Different sequence-to-sequence architectures, such as the Transformer module \cite{vaswani2017attention}, could also be explored and researched more exhaustively. The word-level features provided by FastText could also be replaced by separately trained neural approaches that generate word embeddings from a word's characters (e.g., ELMo embeddings, \citealt{peters2018deep}), and could also be fine-tuned towards specific applications. Another interesting direction includes hybrid word-character architectures, where the encoder receives word-level features, while the decoder operates at the character level. We are also interested in applying our approach to other languages and dialects.

\section*{Acknowledgment}
The second author was supported by the New York University Abu Dhabi Global PhD Student Fellowship program. The support and resources from the High Performance Computing Center at New York University Abu Dhabi are also gratefully acknowledged.

\begin{table}[t]
\setlength{\tabcolsep}{1pt}
\centering
\scalebox{0.94}{
\begin{tabular}{c l c c }
\bf Gold\% & \bf Error Type & \bm{$F_1$} & \bf Example\\ \hline
4.8   & 	Ta Marbuta & 		95.4 & <ktAbh> $\rightarrow$    <ktAbT>\\
29.8 	&	Hamza 	&	92.8 & <tAyiyd> $\rightarrow$    <ta'yiyd> \\
10.5 	&	Space 	&	87.5  & <mAsbab> $\rightarrow$    <mA sabab> \\
0.8 	&	Alif Maqsura	&	83.3 & <AltY> $\rightarrow$    <Alty> \\
0.7 	&	Repeated Letter 	&	81.8 & <AlrjAAAAAl> $\rightarrow$    <AlrjAl> \\
0.6 	&	Wa of Plurality 	&	66.7 & <qAlw> $\rightarrow$    <qAlwA> \\
39.3 	&	Punctuation 	&	56.4 & $\nil$ $\rightarrow$    <.> \\

2.2 	&	Multiple Types 	&	43.1& <'alqyAmah> $\rightarrow$    <AlqyAmaT> \\
1.7 	&	Consonant Switch 	&	41.0 & <^s.h.s> $\rightarrow$    <^sx.s> \\
1.6 	&	Other Types 	&	38.3 & <altql> $\rightarrow$    <alqtl> \\
2.3 	&	Mood \& Case 	&	33.3 & <m.srywn> $\rightarrow$    <m.sryiyn> \\
2.8 	&	Dialect 	&	32.8 & <hyktb> $\rightarrow$    <syktb> \\
1.3 	&	Deleted Letter	&	n/a & <syat.sr> $\rightarrow$    <syant.sr> \\
1.1 	&	Grammar 	&	n/a & <ytajaawz> $\rightarrow$    <ytajaawzwn> \\
0.5 	&	Gold Error	&	n/a & <AlaltY> $\rightarrow$    <AltY> \\

 \hline
\end{tabular}
}
\caption{Error analysis on a sample from the QALB 2014 shared task development set, ordered by $F_1$ score.}
\label{table:errors}
\end{table}

\bibliography{emnlp2018}

\begin{thebibliography}{37}
\expandafter\ifx\csname natexlab\endcsname\relax\def\natexlab#1{#1}\fi

\bibitem[{Abandah et~al.(2015)Abandah, Graves, Al-Shagoor, Arabiyat, Jamour,
  and Al-Taee}]{Abandah2015}
Gheith~A. Abandah, Alex Graves, Balkees Al-Shagoor, Alaa Arabiyat, Fuad Jamour,
  and Majid Al-Taee. 2015.
\newblock Automatic diacritization of {A}rabic text using recurrent neural
  networks.
\newblock \emph{International Journal on Document Analysis and Recognition
  (IJDAR)}, 18(2):183--197.

\bibitem[{Ahmadi(2017)}]{mastersthesis}
Sina Ahmadi. 2017.
\newblock Attention-based encoder-decoder networks for spelling and grammatical
  error correction.
\newblock Master's thesis, Paris Descartes University, 9.

\bibitem[{Bahdanau et~al.(2014)Bahdanau, Cho, and
  Bengio}]{DBLP:journals/corr/BahdanauCB14}
Dzmitry Bahdanau, Kyunghyun Cho, and Yoshua Bengio. 2014.
\newblock Neural machine translation by jointly learning to align and
  translate.
\newblock \emph{CoRR}, abs/1409.0473.

\bibitem[{Belinkov and Glass(2015)}]{belinkov2015arabic}
Yonatan Belinkov and James Glass. 2015.
\newblock Arabic diacritization with recurrent neural networks.
\newblock In \emph{Proceedings of the 2015 Conference on Empirical Methods in
  Natural Language Processing}, pages 2281--2285.

\bibitem[{Bengio et~al.(2015)Bengio, Vinyals, Jaitly, and
  Shazeer}]{DBLP:journals/corr/BengioVJS15}
Samy Bengio, Oriol Vinyals, Navdeep Jaitly, and Noam Shazeer. 2015.
\newblock Scheduled sampling for sequence prediction with recurrent neural
  networks.
\newblock \emph{CoRR}, abs/1506.03099.

\bibitem[{Bojanowski et~al.(2016)Bojanowski, Grave, Joulin, and
  Mikolov}]{bojanowski2016enriching}
Piotr Bojanowski, Edouard Grave, Armand Joulin, and Tomas Mikolov. 2016.
\newblock Enriching word vectors with subword information.
\newblock \emph{arXiv preprint arXiv:1607.04606}.

\bibitem[{Buckwalter(2007)}]{Buckwalter:2007}
Tim Buckwalter. 2007.
\newblock {Issues in Arabic Morphological Analysis}.
\newblock In A.~van~den Bosch and A.~Soudi, editors, \emph{{Arabic
  Computational Morphology: Knowledge-based and Empirical Methods}}. Springer.

\bibitem[{Cho et~al.(2014)Cho, van Merrienboer, Gulcehre, Bahdanau, Bougares,
  Schwenk, and Bengio}]{cho-al-emnlp14}
Kyunghyun Cho, Bart van Merrienboer, Caglar Gulcehre, Dzmitry Bahdanau, Fethi
  Bougares, Holger Schwenk, and Yoshua Bengio. 2014.
\newblock Learning phrase representations using {RNN} encoder--decoder for
  statistical machine translation.
\newblock In \emph{Proceedings of the 2014 Conference on Empirical Methods in
  Natural Language Processing (EMNLP)}, pages 1724--1734, Doha, Qatar.
  Association for Computational Linguistics.

\bibitem[{Dahlmeier and Ng(2012)}]{m2scorer}
Daniel Dahlmeier and Hwee~Tou Ng. 2012.
\newblock A beam-search decoder for grammatical error correction.
\newblock In \emph{Proceedings of the 2012 Joint Conference on Empirical
  Methods in Natural Language Processing and Computational Natural Language
  Learning}, pages 568--578. Association for Computational Linguistics.

\bibitem[{Erdmann et~al.(2018)Erdmann, Zalmout, and
  Habash}]{Erdmann-DAEmbeddings:2018}
Alexander Erdmann, Nasser Zalmout, and Nizar Habash. 2018.
\newblock {Addressing Noise in Multidialectal Word Embeddings}.
\newblock In \emph{Proceedings of the 56th Annual Meeting of the Association
  for Computational Linguistics}.

\bibitem[{Eskander et~al.(2013)Eskander, Habash, Rambow, and
  Tomeh}]{Eskander-Codafy:2013}
Ramy Eskander, Nizar Habash, Owen Rambow, and Nadi Tomeh. 2013.
\newblock {Processing Spontaneous Orthography}.
\newblock In \emph{Proceedings of the 2013 Conference of the North American
  Chapter of the Association for Computational Linguistics: Human Language
  Technologies (NAACL-HLT)}, Atlanta, GA.

\bibitem[{Goldberg(2015)}]{DBLP:journals/corr/Goldberg15c}
Yoav Goldberg. 2015.
\newblock A primer on neural network models for natural language processing.
\newblock \emph{CoRR}, abs/1510.00726.

\bibitem[{Habash et~al.(2012)Habash, Diab, and Rambow}]{Habash_12_LREC}
Nizar Habash, Mona Diab, and Owen Rambow. 2012.
\newblock {Conventional Orthography for Dialectal Arabic}.
\newblock In \emph{Proceedings of the Eighth International Conference on
  Language Resources and Evaluation (LREC-2012)}, pages 711--718, Istanbul,
  Turkey.

\bibitem[{Habash(2010)}]{Habash_10_book}
Nizar~Y Habash. 2010.
\newblock \emph{Introduction to {Arabic} natural language processing},
  volume~3.
\newblock Morgan \& Claypool Publishers.

\bibitem[{Hochreiter and
  Schmidhuber(1997)}]{Hochreiter:1997:LSM:1246443.1246450}
Sepp Hochreiter and J\"{u}rgen Schmidhuber. 1997.
\newblock Long short-term memory.
\newblock \emph{Neural Comput.}, 9(8):1735--1780.

\bibitem[{Ikeda et~al.(2016)Ikeda, Shindo, and Matsumoto}]{ikeda2016japanese}
Taishi Ikeda, Hiroyuki Shindo, and Yuji Matsumoto. 2016.
\newblock Japanese text normalization with encoder-decoder model.
\newblock In \emph{Proceedings of the 2nd Workshop on Noisy User-generated Text
  (WNUT)}, pages 129--137.

\bibitem[{Kingma and Ba(2014)}]{adam}
Diederik~P. Kingma and Jimmy Ba. 2014.
\newblock Adam: {A} method for stochastic optimization.
\newblock \emph{CoRR}, abs/1412.6980.

\bibitem[{Luong et~al.(2015)Luong, Pham, and
  Manning}]{DBLP:journals/corr/LuongPM15}
Minh{-}Thang Luong, Hieu Pham, and Christopher~D. Manning. 2015.
\newblock Effective approaches to attention-based neural machine translation.
\newblock \emph{CoRR}, abs/1508.04025.

\bibitem[{Mikolov et~al.(2013)Mikolov, Sutskever, Chen, Corrado, and
  Dean}]{mikolov2013distributed}
Tomas Mikolov, Ilya Sutskever, Kai Chen, Greg~S Corrado, and Jeff Dean. 2013.
\newblock Distributed representations of words and phrases and their
  compositionality.
\newblock In \emph{Advances in neural information processing systems}, pages
  3111--3119.

\bibitem[{Mohit et~al.(2014)Mohit, Rozovskaya, Habash, Zaghouani, and
  Obeid}]{mohit2014first}
Behrang Mohit, Alla Rozovskaya, Nizar Habash, Wajdi Zaghouani, and Ossama
  Obeid. 2014.
\newblock The first {QALB} shared task on automatic text correction for
  {Arabic}.
\newblock In \emph{Proceedings of the EMNLP 2014 Workshop on {Arabic} Natural
  Language Processing (ANLP)}, pages 39--47.

\bibitem[{Nawar(2015)}]{nawar2015wanlp}
Michael Nawar. 2015.
\newblock {CUFE}$@${QALB}-2015 shared task: {A}rabic error correction system.
\newblock In \emph{Proceedings of the Second Workshop on Arabic Natural
  Language Processing}, pages 133--137, Beijing, China.

\bibitem[{Parker et~al.(2011)Parker, Graff, Chen, Kong, and
  Maeda}]{LDC:Gigaword-5}
Robert Parker, David Graff, Ke~Chen, Junbo Kong, and Kazuaki Maeda. 2011.
\newblock {Arabic Gigaword Fifth Edition}.
\newblock LDC catalog number No. LDC2011T11, ISBN 1-58563-595-2.

\bibitem[{Pasha et~al.(2014)Pasha, Al-Badrashiny, El~Kholy, Eskander, Diab,
  Habash, Pooleery, Rambow, and Roth}]{madamira}
Arfath Pasha, Mohamed Al-Badrashiny, Ahmed El~Kholy, Ramy Eskander, Mona Diab,
  Nizar Habash, Manoj Pooleery, Owen Rambow, and Ryan Roth. 2014.
\newblock {MADAMIRA}: A fast, comprehensive tool for morphological analysis and
  disambiguation of {A}rabic.

\bibitem[{Peters et~al.(2018)Peters, Neumann, Iyyer, Gardner, Clark, Lee, and
  Zettlemoyer}]{peters2018deep}
Matthew~E Peters, Mark Neumann, Mohit Iyyer, Matt Gardner, Christopher Clark,
  Kenton Lee, and Luke Zettlemoyer. 2018.
\newblock Deep contextualized word representations.
\newblock \emph{arXiv preprint arXiv:1802.05365}.

\bibitem[{Rozovskaya et~al.(2015)Rozovskaya, Bouamor, Habash, Zaghouani, Obeid,
  and Mohit}]{rozovskaya2015second}
Alla Rozovskaya, Houda Bouamor, Nizar Habash, Wajdi Zaghouani, Ossama Obeid,
  and Behrang Mohit. 2015.
\newblock {The second QALB shared task on automatic text correction for
  arabic}.
\newblock In \emph{Proceedings of the Second Workshop on {Arabic} Natural
  Language Processing}, pages 26--35.

\bibitem[{Rozovskaya et~al.(2014)Rozovskaya, Habash, Eskander, Farra, and
  Salloum}]{rozovskaya2014thecs}
Alla Rozovskaya, Nizar Habash, Ramy Eskander, Noura Farra, and Wael Salloum.
  2014.
\newblock The {C}olumbia system in the {QALB}-2014 shared task on {Arabic}
  error correction.
\newblock In \emph{ANLP@EMNLP}.

\bibitem[{Srivastava et~al.(2014)Srivastava, Hinton, Krizhevsky, Sutskever, and
  Salakhutdinov}]{srivastava2014dropout}
Nitish Srivastava, Geoffrey Hinton, Alex Krizhevsky, Ilya Sutskever, and Ruslan
  Salakhutdinov. 2014.
\newblock Dropout: a simple way to prevent neural networks from overfitting.
\newblock \emph{The Journal of Machine Learning Research}, 15(1):1929--1958.

\bibitem[{Sutskever et~al.(2014)Sutskever, Vinyals, and
  Le}]{DBLP:journals/corr/SutskeverVL14}
Ilya Sutskever, Oriol Vinyals, and Quoc~V. Le. 2014.
\newblock Sequence to sequence learning with neural networks.
\newblock \emph{CoRR}, abs/1409.3215.

\bibitem[{Trask et~al.(2015)Trask, Gilmore, and Russell}]{trask2015modeling}
Andrew Trask, David Gilmore, and Matthew Russell. 2015.
\newblock Modeling order in neural word embeddings at scale.
\newblock \emph{arXiv preprint arXiv:1506.02338}.

\bibitem[{Vaswani et~al.(2017)Vaswani, Shazeer, Parmar, Uszkoreit, Jones,
  Gomez, Kaiser, and Polosukhin}]{vaswani2017attention}
Ashish Vaswani, Noam Shazeer, Niki Parmar, Jakob Uszkoreit, Llion Jones,
  Aidan~N Gomez, {\L}ukasz Kaiser, and Illia Polosukhin. 2017.
\newblock Attention is all you need.
\newblock In \emph{Advances in Neural Information Processing Systems}, pages
  5998--6008.

\bibitem[{Williams and Zipser(1989)}]{williams1989learning}
Ronald~J Williams and David Zipser. 1989.
\newblock A learning algorithm for continually running fully recurrent neural
  networks.
\newblock \emph{Neural computation}, 1(2):270--280.

\bibitem[{Wu et~al.(2016)Wu, Schuster, Chen, Le, Norouzi, Macherey, Krikun,
  Cao, Gao, Macherey, Klingner, Shah, Johnson, Liu, Łukasz Kaiser, Gouws,
  Kato, Kudo, Kazawa, Stevens, Kurian, Patil, Wang, Young, Smith, Riesa,
  Rudnick, Vinyals, Corrado, Hughes, and Dean}]{45610}
Yonghui Wu, Mike Schuster, Zhifeng Chen, Quoc~V. Le, Mohammad Norouzi, Wolfgang
  Macherey, Maxim Krikun, Yuan Cao, Qin Gao, Klaus Macherey, Jeff Klingner,
  Apurva Shah, Melvin Johnson, Xiaobing Liu, Łukasz Kaiser, Stephan Gouws,
  Yoshikiyo Kato, Taku Kudo, Hideto Kazawa, Keith Stevens, George Kurian,
  Nishant Patil, Wei Wang, Cliff Young, Jason Smith, Jason Riesa, Alex Rudnick,
  Oriol Vinyals, Greg Corrado, Macduff Hughes, and Jeffrey Dean. 2016.
\newblock Google's neural machine translation system: Bridging the gap between
  human and machine translation.
\newblock \emph{CoRR}, abs/1609.08144.

\bibitem[{Xie et~al.(2016)Xie, Avati, Arivazhagan, Jurafsky, and
  Ng}]{Xie2016NeuralLC}
Ziang Xie, Anand Avati, Naveen Arivazhagan, Daniel Jurafsky, and Andrew~Y. Ng.
  2016.
\newblock Neural language correction with character-based attention.
\newblock \emph{CoRR}, abs/1603.09727.

\bibitem[{Yuan and Briscoe(2016)}]{yuan2016grammatical}
Zheng Yuan and Ted Briscoe. 2016.
\newblock Grammatical error correction using neural machine translation.
\newblock In \emph{Proceedings of the 2016 Conference of the North American
  Chapter of the Association for Computational Linguistics: Human Language
  Technologies}, pages 380--386.

\bibitem[{Zaghouani et~al.(2014)Zaghouani, Mohit, Habash, Obeid, Tomeh,
  Rozovskaya, Farra, Alkuhlani, and Oflazer}]{zaghouani2014large}
Wajdi Zaghouani, Behrang Mohit, Nizar Habash, Ossama Obeid, Nadi Tomeh, Alla
  Rozovskaya, Noura Farra, Sarah Alkuhlani, and Kemal Oflazer. 2014.
\newblock Large scale {Arabic} error annotation: Guidelines and framework.
\newblock In \emph{International Conference on Language Resources and
  Evaluation (LREC 2014)}.

\bibitem[{Zalmout et~al.(2018)Zalmout, Erdmann, and
  Habash}]{Zalmout-EGYMorphology:2018}
Nasser Zalmout, Alexander Erdmann, and Nizar Habash. 2018.
\newblock {Noise-Robust Morphological Disambiguation for Dialectal Arabic}.
\newblock In \emph{Proceedings of the 2018 Conference of the North American
  Chapter of the Association for Computational Linguistics: Human Language
  Technologies (NAACL-HLT)}.

\bibitem[{Zalmout and Habash(2017)}]{Nasser2017EMNLP}
Nasser Zalmout and Nizar Habash. 2017.
\newblock Don't throw those morphological analyzers away just yet: Neural
  morphological disambiguation for {Arabic}.
\newblock In \emph{Proceedings of the 2017 Conference on Empirical Methods in
  Natural Language Processing}, pages 704--713, Copenhagen, Denmark.
  Association for Computational Linguistics.

\end{thebibliography}
\bibliographystyle{acl_natbib_nourl}

\end{document}